\documentclass[11pt]{article}
\usepackage[margin=1in]{geometry}
\usepackage{amsmath,amssymb,amsthm}
\usepackage{graphicx}
\usepackage{booktabs}
\usepackage{microtype}
\usepackage[numbers,sort&compress]{natbib}
\usepackage[colorlinks=true,linkcolor=blue,citecolor=blue,urlcolor=blue]{hyperref}
\usepackage{xcolor}

\newcommand{\hstar}{h^{*}}
\newcommand{\zzero}{z_{0}}

\title{The Equilibrium Is the Initialization:\\
Lazy Identity Collapse in Physics-Structured Deep Equilibrium Reasoning}
\author{Joyjeet Singh\\
Independent Researcher\\
\texttt{ORCID: 0009-0005-1512-7439}}
\date{\today}

\begin{document}
\maketitle

\begin{abstract}
Deep equilibrium models promise input-adaptive implicit computation:
harder problems should demand more solver iterations, and the solved
equilibrium should encode the result of genuine iterative inference.
We report a cautionary study of a port-Hamiltonian DEQ with a learned
initialization on two reasoning tasks --- ProofWriter entailment over
frozen DeBERTa embeddings and a BFS-verified graph-reachability
benchmark --- in which the implicit computation is a silent no-op.
Across tasks, seeds, and controlled ablation arms, the solved
equilibrium equals the solver's \emph{start point} to numerical
precision, and bypassing the solver entirely changes test accuracy by
$+0.00$ percentage points in 18 of 19 training runs. Controlled
interventions falsify the tempting explanation: removing the anchoring
term reproduces every result, and retraining with noise-decoupled
starts yields a solver that converges to the noisy start while the
decoder learns to ignore it. The single escaping run diverges instead
($\lVert h^{*}{-}z_0\rVert = 171$), producing a co-adapted noise
channel whose removal \emph{improves} accuracy. Iteration counts are
uncorrelated with ground-truth difficulty ($r=0.009$), and the full
apparatus never outperforms a two-layer MLP on either task. We trace
the mechanism to gradient starvation along two distinct routes, show
that the standard zeroing ablation is confounded and gives wildly
seed-dependent answers where the correct substitution test gives a
stable zero, and distill a four-test diagnostic protocol for auditing
claimed implicit computation. All experiments run on a single free
Colab GPU; code, raw logs, and analysis scripts are released.
\end{abstract}

\section{Introduction}\label{sec:intro}
Implicit and recurrent depth is one of the field's recurring hopes for
reasoning: rather than fixing computation at a static layer count, let
the model iterate --- a deep equilibrium model (DEQ) solving for a
fixed point~\citep{bai2019deep}, a looped or recurrent-depth
transformer unrolling until done~\citep{dehghani2019universal,
geiping2025scaling} --- and let harder inputs consume more
computation~\citep{graves2016adaptive}.
Physics-structured parameterizations make the proposal more
attractive still: port-Hamiltonian and symplectic constraints promise
stable, energy-controlled latent dynamics on which an implicit solve
can safely iterate~\citep{greydanus2019hamiltonian, jin2020sympnets,
desai2021port}.
The promise is concrete and checkable: the solved equilibrium should
carry the \emph{result} of iterative inference, and solver effort
should track problem difficulty.

This paper documents, in controlled detail, a design in which every
part of that promise fails silently. We study a port-Hamiltonian DEQ
whose state is initialized by a learned, goal-conditioned feed-forward
network --- a natural pattern whenever an implicit layer is grafted onto
a pretrained encoder, and one that (with or without an explicit
anchoring term pulling the solve toward the initialization) appears in
many variants across the implicit-models literature. Training
converges, accuracies look respectable, and solver telemetry looks
plausible. Yet the equilibrium is a copy of its initialization: across
two tasks, nineteen training runs, an anchor-strength sweep, and three
intervention arms, the fixed-point solve moves the state by at most
$10^{-6}$ and contributes exactly $+0.00$ percentage points of
accuracy, while iteration counts ignore a ground-truth difficulty
signal entirely.

The obvious mechanistic story --- the anchor both starts and attracts
the solver, so the dynamics never receive off-anchor gradients --- turns
out to be wrong, and we consider its falsification the most useful
part of the study. Removing the anchor changes nothing; retraining
with noise-decoupled starts produces a solver that faithfully
converges to the \emph{noisy} start while the decoder learns to ignore
the channel; and the one run that escapes the identity regime does so
by diverging, yielding a co-adapted noise channel whose removal
improves accuracy. The correct diagnosis is gradient starvation: both
tasks are solvable through the direct context pathway --- a two-layer
MLP baseline matches or beats the full apparatus throughout --- so
``compute nothing'' is a stable optimum, reached by two different
gradient routes in our two pipelines. Along the way we show that the
standard zeroing ablation is confounded for implicit modules,
producing seed-dependent deltas as large as $+44$pp where the correct
substitution test gives a stable zero.

Our contributions: (i) a mechanistically dissected case study of
silent no-op implicit computation, with the anchor-causation
hypothesis explicitly tested and falsified by controlled retraining
arms; (ii) identification of two degenerate regimes --- lazy identity
collapse and solver blow-up --- with per-seed evidence that pooled
statistics would obscure; (iii) a demonstration that zeroing ablations
confound pathway with computation, and the substitution test that
deconfounds them; (iv) a four-test diagnostic protocol
(Section~\ref{sec:protocol}) cheap enough to run on a free GPU and
applicable to DEQs, looped, and recurrent-depth models generally; and
(v) a fully reproducible negative result: every number traces to
released raw logs via a released analysis script.

\section{Related Work}\label{sec:related}
\paragraph{Deep equilibrium models and implicit layers.}
Deep equilibrium models compute their output as the fixed point of a
learned layer, differentiated implicitly~\citep{bai2019deep,
bai2020multiscale, elghaoui2021implicit}, with guarantees available
under monotonicity assumptions~\citep{winston2020monotone} and
practical solves typically using Anderson
acceleration~\citep{anderson1965}. Known DEQ pathologies concentrate
on the \emph{unstable} side: growing iteration counts and brittle
convergence, addressed by Jacobian
regularization~\citep{bai2021stabilizing}, by cheaper approximate
gradients~\citep{fung2022jfb}, and by encouraging path
independence, which correlates with the ability to exploit extra
test-time iterations~\citep{anil2022path}. Our study documents the
opposite, quieter failure: \emph{degenerate convergence}, in which the
solve succeeds trivially because the equilibrium is the
initialization. Path independence offers a useful lens here --- a model
whose output ignores initialization because it computes something is
the desirable extreme; ours ignores computation because it copies the
initialization.

\paragraph{Physics-structured latent dynamics.}
Encoding mechanics into network dynamics --- Hamiltonian neural
networks~\citep{greydanus2019hamiltonian}, Lagrangian
variants~\citep{cranmer2020lagrangian}, symplectic
parameterizations~\citep{jin2020sympnets}, and port-Hamiltonian
extensions with explicit dissipation~\citep{desai2021port}, in the
broader family of continuous-depth models~\citep{chen2018neural} ---
promises stable, energy-controlled latent evolution. Our dynamics use
the standard structure--dissipation decomposition with learned kinetic
and potential energies and a context-conditioned damping term. The
irony our results add: the same properties that make such dynamics
stable to integrate make ``stay where you started'' an easily
reachable, easily maintained solution when nothing in the loss demands
motion.

\paragraph{Recurrent depth, looped models, and test-time compute.}
Allocating more sequential computation to harder inputs is an old and
newly urgent idea: adaptive computation time~\citep{graves2016adaptive,
banino2021pondernet}, universal
transformers~\citep{dehghani2019universal}, recurrent networks that
extrapolate by iterating longer~\citep{schwarzschild2021can,
bansal2022end}, looped transformers~\citep{yang2024looped},
recurrent-depth language models~\citep{geiping2025scaling}, and tiny
recursive reasoners~\citep{wang2025hierarchical, jolicoeur2025less},
evaluated on algorithmic suites such as
CLRS~\citep{velickovic2022clrs}. Claims in this literature share a
load-bearing assumption: that the iterated module actually computes.
Our protocol (Section~\ref{sec:protocol}) is a cheap audit of exactly
that assumption, and our case study shows the assumption can fail
silently while every surface metric looks healthy.

\paragraph{Negative results and diagnostics.}
Our probe methodology follows the linear-probing
tradition~\citep{alain2017understanding} with the caution urged by
control tasks~\citep{hewitt2019designing} --- our blow-up regime is a
concrete instance of a probe finding ``signal'' that is not
computation. The mechanism we identify is a module-level relative of
gradient starvation~\citep{pezeshki2021gradient} and of shortcut
learning~\citep{geirhos2020shortcut}: a direct pathway that suffices
for the loss removes the incentive for the expensive pathway to learn.
We use the term in this architectural sense rather than the
feature-dynamics sense of~\citet{pezeshki2021gradient}.

\section{Model and Tasks}\label{sec:setup}
\subsection{Port-Hamiltonian anchored DEQ}
The implicit state is phase-space--structured, $z=(q,p)$, evolved by a
damped, anchored, symplectic-Euler-style update conditioned on a
context vector $c$:
\begin{align}
q_{t+1} &= q_t + \epsilon\,\nabla_p K(p_t, c) - \epsilon\, r(c)\, p_t
          + \epsilon\, r_a (q_0 - q_t),\label{eq:qstep}\\
p_{t+1} &= p_t - \epsilon\,\nabla_q U(q_{t+1}, c) - \epsilon\, r(c)\, q_{t+1}
          + \epsilon\, r_a (p_0 - p_t),\label{eq:pstep}
\end{align}
where $K$ and $U$ are learned scalar (kinetic and potential) MLPs,
$r(c)=\mathrm{softplus}(w_r^\top c + b_r)$ is a learned
context-conditioned damping coefficient, $\epsilon=0.05$, and
$r_a$ is the anchor strength ($0.02$ by default; $0$ in arm C) pulling
the state toward the learned initialization
$\zzero=(q_0,p_0)$. Weight matrices of $K$ and $U$ are spectrally
clamped to norm $\le 0.95$ after every update. The fixed point of
Eqs.~(\ref{eq:qstep})--(\ref{eq:pstep}) is solved by Anderson
acceleration ($m{=}5$, cap $300$ iterations) from a start point that
equals $\zzero$ in the standard design, to a per-example adaptive
tolerance $10^{-(5\sigma(w_t^\top c + b_t)+1)}\in[10^{-6},10^{-1}]$.
Gradients pass through the implicit function theorem via an adjoint
Anderson solve. The initialization network maps context to phase
space: $q_0$ by a linear read-out of the premise/graph context, and
$p_0$ through a goal-gated channel in which a query--key gate modulates
value projections of the context; we study both a scalar-gate and an
elementwise-gate variant of this channel
(Section~\ref{sec:gate}). The decoder is a two-layer GELU MLP reading
$[\hstar; c_{\text{ctx}}; c_{\text{query}}]$. Backward-pass coverage
of $\zzero$ differs between our two pipelines and matters for the
mechanism; see Section~\ref{sec:starvation}.

\subsection{Tasks and positive controls}
\textbf{ProofWriter entailment.} Binary entailment on
ProofWriter~\citep{tafjord2021proofwriter}, stratified over proof
depths $\{1,2,3,5\}$ ($800$ per depth where available; depth-5
capped at $697$), yielding $2477$ train / $620$ test examples with
label balance $0.495/0.477$. Premises and goals are embedded by frozen
DeBERTa-v3-base~\citep{he2023debertav3}; embeddings are cached, so
every method reads identical inputs. \textbf{Graph reachability.}
Synthetic $k$-hop reachability: random directed graphs on $16$ nodes
with out-degree $2$; a query $(s,t,k)$ asks whether $t$ is reachable
from $s$ within $k\le 8$ hops, labeled by breadth-first search and
balanced exactly ($6{,}000$ train / $1{,}000$ test; $30{,}000$ train
in arm E). Every instance carries its true BFS distance, giving a
ground-truth difficulty signal for the iteration analysis
(Section~\ref{sec:iters}). \textbf{Positive controls.} A two-layer
MLP on the identical cached inputs is trained fresh per seed alongside
every pipeline run ($90.39\pm0.33\%$ on ProofWriter; the analogous
context-only model is arm D on graphs), establishing both that the
data pipeline supports learning and what the implicit apparatus must
beat.\footnote{Working notes from an unpreserved earlier session
recorded an $81.72\%$ baseline; it failed re-verification against
executed outputs and is superseded by the fresh per-seed values
reported here (see Limitations, reproducibility discipline).}

\section{Results: Two Degenerate Regimes}\label{sec:collapse}

\subsection{The equilibrium is the start point}\label{sec:lazy}
\begin{figure}[t]\centering
\includegraphics[width=.72\linewidth]{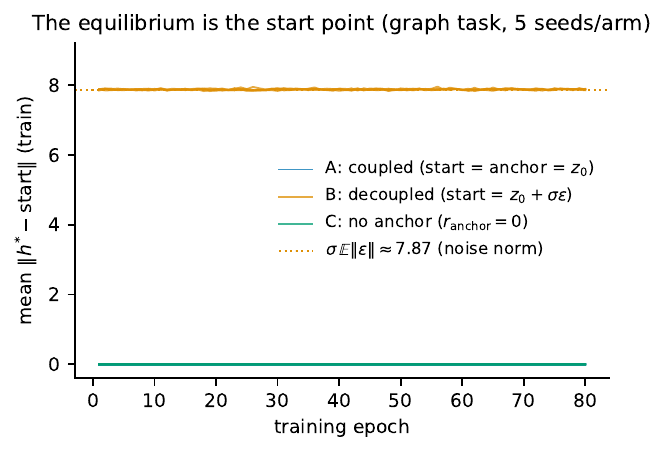}
\caption{Training-time displacement between the solved equilibrium and
the solver's start point, graph task, all seeds. Arms A (coupled) and C
(anchor removed) sit at zero; arm B (retrained from noisy starts) sits
exactly on the injected-noise norm. The equilibrium is the
initialization.}\label{fig:lazy}
\end{figure}

Figure~\ref{fig:lazy} shows the training-time displacement
$\lVert\hstar-\text{start}\rVert$ for every seed of three controlled
arms on the graph task. In arm~A (the original design: solver started
at the anchor $\zzero$), displacement is numerically zero throughout
training and remains so at test time (mean $6.9\times10^{-7}$, pooled
over 5 seeds). Removing the anchor entirely (arm~C, $r_\text{anchor}=0$)
changes nothing: per-seed accuracies are identical to arm~A and loss
curves agree to four decimal places, with displacement differing only
at the seventh decimal. The anchoring term is causally irrelevant to
the collapse. Most tellingly, when the solver is \emph{retrained} from
noisy starts $\zzero+\sigma\varepsilon$ (arm~B, $\sigma{=}2$,
$\varepsilon$ resampled every forward pass), training displacement sits
throughout on the theoretical norm of the injected noise,
$\sigma\,\mathbb{E}\lVert\varepsilon\rVert\approx7.87$: the solved
``equilibrium'' is the start point, wherever the start is placed. The
same invariance holds across a five-value sweep of the anchor strength
$r_\text{anchor}\in\{0.005,0.02,0.05,0.1,0.2\}$ (displacement
$2\times10^{-6}$ at every value) and persists when the displacement
penalty is removed from the loss, ruling out the regularizer as the
cause. On ProofWriter, 4 of 5 seeds show the same signature
(displacement $\le0.12$; Table~\ref{tab:pw}); the fifth exhibits the
second regime, described in Section~\ref{sec:blowup}.

\subsection{Bypassing the solver changes nothing}\label{sec:subst}
The unconfounded measure of the DEQ's contribution is the
\emph{substitution test}: feed the solver's start point directly to
the decoder, skipping the fixed-point solve entirely. Table~\ref{tab:arms}
reports the result on the graph task: the gap between the full pipeline
and the solver-bypassed pipeline is $+0.00$pp in every arm and every
seed --- including arm~B, which was explicitly retrained to break the
start--anchor coupling, and arm~E, which scales training data
$5\times$. A context-only baseline (arm~D: identical embedding and
decoder, no initializer, no DEQ) matches the full pipeline
($61.52\pm0.72$ vs.\ $61.80\pm1.34$), so the \emph{entire} implicit
apparatus is dispensable. On ProofWriter the substitution gap is
$\le0.32$pp in the 4 collapsed seeds (Table~\ref{tab:pw}).

\begin{table}[t]\centering\small
\caption{Graph reachability, mean$\pm$std over seeds. ``Start-sub''
bypasses the solver, feeding its start point to the decoder. The
implicit computation contributes $+0.00$pp everywhere; the context-only
arm matches the full pipeline. All arms reach 100\% train accuracy.}
\label{tab:arms}
\begin{tabular}{lccccc}\toprule
Arm & Full & Start-sub & Gap (pp) & $\hstar$-zeroed & $\lVert\hstar-\zzero\rVert$\\\midrule
A coupled (5s) & $61.80\pm1.34$ & $61.80\pm1.34$ & $0.00\pm0.00$ & $60.92\pm1.01$ & $7{\times}10^{-7}$\\
B decoupled retrain (5s) & $61.82\pm0.96$ & $61.82\pm0.96$ & $0.00\pm0.00$ & $61.84\pm0.96$ & $3{\times}10^{-7}$\\
C no anchor (5s) & $61.80\pm1.34$ & $61.80\pm1.34$ & $0.00\pm0.00$ & $60.92\pm1.01$ & $7{\times}10^{-7}$\\
E $30$k graphs (3s) & $64.20\pm0.79$ & $64.20\pm0.79$ & $0.00\pm0.00$ & $64.37\pm1.32$ & $7{\times}10^{-7}$\\\midrule
D context-only (5s) & $61.52\pm0.72$ & --- & --- & --- & ---\\
\bottomrule\end{tabular}\end{table}

\begin{table}[t]\centering\small
\caption{ProofWriter, per seed (pooling would average two distinct
regimes). Seeds 1, 2, 4, 5: lazy identity collapse. Seed 3: solver
blow-up --- start-substitution falls below chance while \emph{zeroing}
$\hstar$ improves over the full model. The fresh per-seed MLP baseline
(right) is never beaten.}\label{tab:pw}
\begin{tabular}{lcccccc}\toprule
Seed & Full & Start-sub & Gap (pp) & $\hstar$-zeroed & $\lVert\hstar-\zzero\rVert$ & MLP baseline\\\midrule
1 & 91.77 & 91.77 & $+0.00$ & 47.74 & $2{\times}10^{-6}$ & 90.65\\
2 & 91.29 & 91.61 & $-0.32$ & 53.39 & $0.12$ & 90.32\\
3 & 81.77 & 32.58 & $+49.19$ & \textbf{83.23} & $\mathbf{171.43}$ & 90.00\\
4 & 90.32 & 90.32 & $+0.00$ & 47.74 & $2{\times}10^{-6}$ & 90.81\\
5 & 91.77 & 91.77 & $+0.00$ & 52.26 & $1{\times}10^{-7}$ & 90.16\\
\bottomrule\end{tabular}\end{table}

\begin{figure}[t]\centering
\includegraphics[width=.95\linewidth]{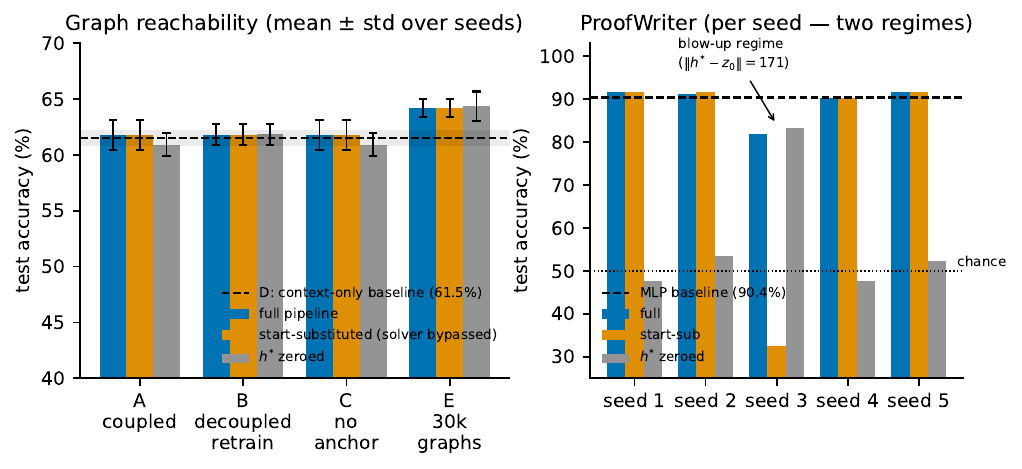}
\caption{Substitution and ablation accuracies. Left: graph task
(mean$\pm$std); the dashed line is the context-only arm~D. Right:
ProofWriter per seed; seed 3 is the blow-up regime.}\label{fig:subst}
\end{figure}

\subsection{Iteration counts ignore difficulty}\label{sec:iters}
\begin{figure}[t]\centering
\includegraphics[width=.95\linewidth]{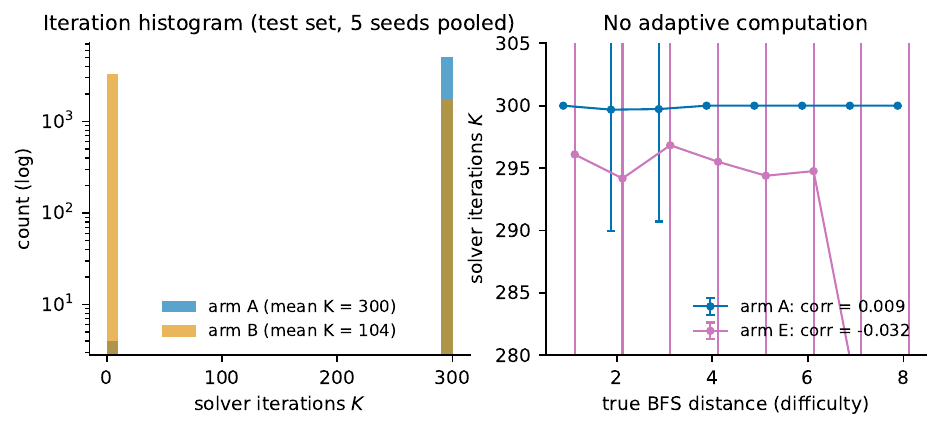}
\caption{Left: iteration histograms (arm B's bimodality reflects noisy
starts that occasionally satisfy the tolerance immediately). Right:
mean $K$ vs.\ ground-truth BFS distance; an adaptive solver would slope
upward.}\label{fig:iters}
\end{figure}

If the implicit layer performed input-adaptive computation, harder
instances should require more solver iterations. They do not
(Figure~\ref{fig:iters}). On the graph task --- where every instance
carries a BFS-verified ground-truth distance --- the correlation
between iteration count $K$ and true distance is $0.009$ (arm~A,
5 seeds pooled; $-0.032$ at $5\times$ data), with $99.9\%$ of solves
running to the iteration cap regardless of difficulty. Accuracy is
likewise constant across solver tolerances from $10^{-2}$ to
$10^{-6}$: since the state never moves, solve precision is irrelevant.

\subsection{The minority regime: solver blow-up}\label{sec:blowup}
\begin{figure}[t]\centering
\includegraphics[width=.6\linewidth]{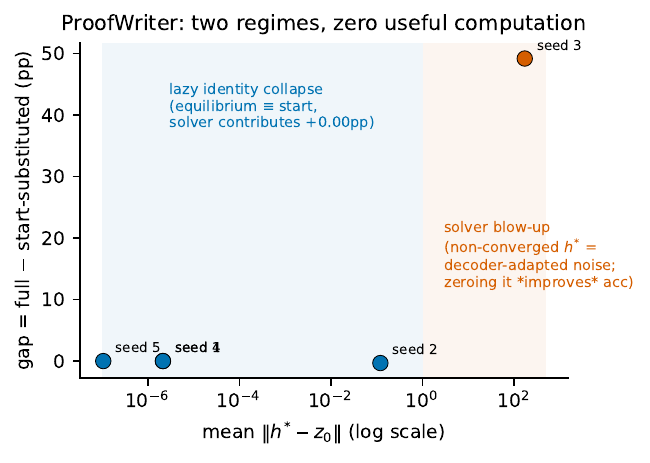}
\caption{ProofWriter seeds in the (displacement, substitution-gap)
plane. Four seeds collapse to the start; one diverges. Neither regime
performs useful implicit computation.}\label{fig:regimes}
\end{figure}

One ProofWriter seed in five escapes the identity collapse --- by
diverging. Seed 3 ends training with mean displacement $171.4$; its
start-substituted accuracy falls to $32.58\%$, \emph{below} the $47.7\%$
majority-class floor, while zeroing $\hstar$ \emph{raises} accuracy
from $81.77\%$ to $83.23\%$ (Table~\ref{tab:pw},
Figure~\ref{fig:regimes}). The interpretation is unflattering to the
implicit layer in a different way: the non-converged output acts as a
high-variance channel the decoder has co-adapted to, and removing it
outright helps. We observed the same instability earlier when an
elementwise initializer gate increased gradient flow by an order of
magnitude: solver residuals grew from $0.09$ to a mean of $29.1$
(max $48.3$) while a probe on $\hstar$ appeared to find signal --- signal
riding on divergence, not equilibrium (Appendix~\ref{app:curves}). The
two regimes jointly close the escape routes: the equilibrium either
stays at its start (and contributes nothing) or leaves it by diverging
(and contributes noise).

\subsection{The apparatus never beats a two-layer MLP}\label{sec:baseline}
Finally, the positive controls double as an upper-bound check. On
ProofWriter, a two-layer MLP on the same frozen embeddings reaches
$90.39\pm0.33\%$ (fresh initialization per seed); the full
physics-structured pipeline reaches $89.39\pm4.30\%$. On the graph
task, the context-only arm matches the full pipeline. Across both
tasks, at no point does the port-Hamiltonian DEQ apparatus outperform
the simplest baseline it was meant to improve upon.

\section{Mechanism}\label{sec:mechanism}

\subsection{Where the signal actually flows}\label{sec:signal}
On ProofWriter, all task-relevant information travels through the
feed-forward initializer's goal-gated momentum channel and the
decoder's direct view of the context. Zeroing the $p_0$ (goal-gated)
half of the initializer output at inference drops accuracy from
$91.77\%$ to $47.74\%$ --- the majority-class floor --- while zeroing
the $q_0$ half changes nothing ($91.77\%$). Accuracy from the start
point alone is essentially flat in proof depth ($92.5/92.5/91.9/90.0\%$
at depths $1/2/3/5$), consistent with a task solvable by a shallow
feed-forward read-out rather than iterative inference. A fresh
classifier probing the frozen equilibrium $\hstar$ \emph{alone}
converges to the label-entropy floor (loss $\approx\ln 2$; test
accuracy $51.45\%$): the equilibrium carries no task signal beyond what
it copies from its start.

\subsection{Causal dissection: what does and does not cause the collapse}
\label{sec:causal}
The design invites a seductive mechanistic story: the anchor both
initializes the solver and attracts it, so the solver starts where it
is pulled back to, displacement is identically zero, and the
Hamiltonian networks never receive gradients away from the anchor --- a
self-fulfilling no-op. Two controlled arms falsify the causal part of
this story. Removing the anchor entirely (arm~C) reproduces arm~A's
per-seed results to reported precision. Retraining with the start
decoupled from the anchor (arm~B) --- the intervention the
self-fulfilling-no-op account predicts should rescue the dynamics ---
does not: the solver remains a near-identity map on its (now noisy)
start, and the decoder's adaptation is to \emph{ignore} the $\hstar$
channel, as witnessed by its zeroing ablation matching full accuracy
exactly ($61.84\pm0.96$ vs.\ $61.82\pm0.96$). Forcing displaced starts
at inference on the trained coupled model completes the picture:
displacement becomes genuine ($\approx7.9$) but accuracy \emph{drops}
($56.06\pm1.26$ vs.\ $61.80\pm1.34$), so the off-start regions of the
learned vector field contain no structure --- capability absent, not
suppressed.

\subsection{Gradient starvation: two routes to the same no-op}
\label{sec:starvation}
The deeper cause is that nothing in the gradient structure ever
rewards genuine iterative computation, and a near-identity fixed-point
map is a stable optimum. Both tasks are solvable from the direct
context pathway (Section~\ref{sec:baseline}), so the loss can be
driven to zero while the equilibrium contributes nothing; once it is,
the gradient reaching the dynamics networks through the implicit
function theorem is proportional to a decoder sensitivity that the
decoder itself is free to remove. The two pipelines realize this
starvation by different routes, which is worth stating precisely
because it shows the phenomenon is not an implementation quirk.
In the ProofWriter pipeline, the backward pass extends to $\zzero$, the
initializer trains, and $\zzero$ genuinely carries signal
(Section~\ref{sec:signal}) --- but the solve on top of it is an
identity. In the graph pipeline, the custom VJP returns zero
cotangents for both $\zzero$ and the context, so the initializer is
\emph{gradient-isolated} and remains at its random initialization;
the decoder learns from its direct context inputs, and the equilibrium
faithfully copies an untrained random projection. Same symptom, two
gradient-starvation routes: one starves the dynamics of incentive, the
other additionally starves the start of training. At initialization
the port-Hamiltonian update is already near-identity (small MLP
outputs, step size $\epsilon=0.05$, spectral norms clamped below
$0.95$), so ``stay put'' is where training begins --- and nothing ever
pays it to leave, except, occasionally, instability
(Section~\ref{sec:blowup}).

\section{A Note on Confounded Ablations}\label{sec:confound}
Our own first attempt at measuring the DEQ's contribution was wrong in
an instructive way. The natural ablation --- zero the equilibrium
$\hstar$ at the decoder input --- produced wildly seed-dependent
deltas on ProofWriter ($+44.0, +19.5, +0.0, +44.0, +39.5$\,pp across
five seeds), which superficially read as ``the DEQ sometimes matters
a great deal.'' It does not. Zeroing an internal representation
removes \emph{two} things at once: the computation performed on the
representation, and the input pathway flowing \emph{through} it ---
here, the start point $\zzero$ that the identity-solve copies to the
decoder. The seed variance in the zeroing deltas tracks how each
decoder happens to distribute weight between its $\hstar$ channel and
its direct context channels, not what the solver computes. The
substitution test isolates the computation by keeping the pathway and
bypassing only the solve, and it returns $0.00$\,pp with negligible
variance in every collapsed run. The blow-up seed makes the confound
vivid from the other side: there, zeroing \emph{helps}
(Section~\ref{sec:blowup}), because the ``pathway'' being removed is
co-adapted noise. We suggest the general rule: whenever an iterative
or implicit module is credited with computation, the control should
substitute the module's input for its output --- not delete the
channel.

\section{A Diagnostic Protocol for Implicit Computation}\label{sec:protocol}
The failure we document is silent: training converges, accuracy is
respectable, solver telemetry looks plausible, and the model card
would read ``port-Hamiltonian deep equilibrium reasoner.'' Nothing in
the standard training signals reveals that the equilibrium computation
is a no-op. We therefore distill our analysis into four inexpensive
tests, applicable to any architecture that claims iterative or
implicit computation over a learned initialization --- DEQs, looped
transformers, and recurrent-depth models alike. On our tasks the full
protocol runs in minutes on a free Colab GPU.

\paragraph{T1 --- Substitution (the primary test).} At evaluation,
replace the module's output with its input: feed the solver's start
point (or the loop's initial state) directly to the decoder.
\emph{Fail} if accuracy is unchanged (the module computes nothing);
in our runs the gap was $+0.00$\,pp across 18 of 19 training runs
spanning two tasks. This test --- not the zeroing ablation --- isolates
the computation from the pathway (Section~\ref{sec:confound}).

\paragraph{T2 --- Representation probe against the entropy floor.}
Train a fresh classifier on the module's output \emph{alone}, frozen.
\emph{Fail} if probe loss converges to the label-entropy floor
($\ln 2$ for balanced binary tasks; $51.45\%$ accuracy in our case).
A passing probe with a \emph{diverging} solver should be treated as a
false positive: verify residuals before crediting the representation
(our gate-fix run probed at $88.9\%$ while mean residuals sat at
$29.1$).

\paragraph{T3 --- Iteration variance against difficulty.} On a task
with a controllable or verifiable difficulty measure (here,
ground-truth BFS distance), correlate solver iterations with
difficulty. \emph{Fail} if the correlation is
$\approx 0$ or iteration counts are degenerate (pinned at the cap, or
at immediate exit): the model is not allocating computation to
instances. Our runs: corr $= 0.009$ with $99.9\%$ of solves at the
cap.

\paragraph{T4 --- Forced displacement.} Start the solver away from its
learned initialization at inference (we used Gaussian offsets with
$\sigma$ matched to the state scale). \emph{Fail-absent} if accuracy
degrades with genuine displacement --- the off-initialization vector
field was never trained and contains no structure --- as opposed to
\emph{fail-suppressed}, where accuracy is maintained, indicating a
capability the training regime masks. Our runs degrade
($61.8\%\!\to\!56.1\%$), and the retraining version of this test
(arm B) shows the decoder simply learns to ignore the module.

A model that passes all four tests may still not ``reason,'' but a
model that fails T1 certainly does not, whatever its architecture
diagram promises. We suggest T1 as a standard reported control
whenever implicit or recurrent computation is part of a paper's
claimed contribution.

\section{Secondary Finding: Expressive Gates Destabilize the Solve}\label{sec:gate}
During the ProofWriter study we replaced the scalar goal gate in the
initialization network with an elementwise gate, which increased the
gradient norm reaching the initializer by an order of magnitude
($0.0027\to0.026$) --- and destabilized the solve: mean final
residuals grew from $0.09$ to $29.1$ (max $48.3$), so the returned
$\hstar$ was no longer an equilibrium at all. Instructively, a probe
on this non-converged $\hstar$ reached $88.9\%$: apparent
representational ``signal'' riding entirely on divergence, foreshadowing
the seed-3 blow-up regime of Section~\ref{sec:blowup}. An instrumented
run captured the onset window (epochs 74--90): peak state magnitude
grew $6.9\to21.7$, initializer gradients spiked to $1.7$, and
accuracy fell from $92\%$ to $62\%$ before partially recovering
(Appendix~\ref{app:curves}). The general lesson is a trade-off worth
naming: interventions that increase gradient flow into the
initialization of an implicit model can buy trainability at the price
of solver validity --- and standard telemetry (accuracy, probe scores)
does not distinguish the two.

\section{Limitations}\label{sec:limitations}
\paragraph{Scope of the negative claim.} Our evidence concerns one
architecture family (port-Hamiltonian dynamics with a learned
initializer, solved by Anderson acceleration with $m{=}5$ and a
300-iteration cap) on two binary classification tasks. We claim that
\emph{this} design performs no useful implicit computation and that
the mechanism is gradient starvation; we do not claim that all DEQs,
looped models, or physics-structured implicit layers must fail.
Indeed, the diagnostic framing exists precisely because the failure is
silent and design-specific. Tasks where the direct context pathway
\emph{cannot} solve the problem --- unlike both of ours
(Section~\ref{sec:baseline}) --- may exert the gradient pressure our
setup lacks; establishing this is future work.

\paragraph{The two pipelines differ in backward-pass coverage.} In the
ProofWriter pipeline the backward pass extends to the initializer; in
the graph pipeline the custom VJP returns zero cotangents for
$\zzero$ and the context, leaving the initializer untrained
(Section~\ref{sec:starvation}). We report this as two routes to the
same starvation, but it means the fully-differentiated configuration
was tested on only one of the two tasks. The substitution result is
unaffected --- it is a forward-pass property and holds identically in
both pipelines.

\paragraph{Generalization on the synthetic task.} All graph-task arms
memorize (100\% train accuracy) and generalize weakly ($\sim$62\% at
6{,}000 graphs, $64.2\%$ at 30{,}000). The lazy-identity result is
orthogonal to this gap --- the substitution test compares two decoders
reading the same representations --- but task-level accuracies on the
graph benchmark should not be read as statements about reachability
reasoning in general.

\paragraph{The blow-up regime is characterized from few events.} The
divergent regime appears in 1 of 5 ProofWriter seeds (plus,
mechanistically consistent, in our earlier gate-modified run). We can
assert its existence and its signature --- below-chance substitution,
zeroing that improves accuracy, displacement orders of magnitude above
the collapsed regime --- but not its frequency as a function of
hyperparameters.

\paragraph{Frozen encoders.} Both tasks use frozen or from-scratch
context encoders (DeBERTa-v3-base embeddings; a small trained
embedding for graphs). End-to-end finetuning of a large encoder could
change the gradient landscape reaching the implicit layer.

\paragraph{Reproducibility discipline.} All numbers in this paper are
recomputed from raw per-run logs by a released analysis script; two
figures quoted in our own working notes from an unpreserved earlier
session (a $81.72\%$ baseline; a $51.29\%$ probe) were retired when
they failed re-verification, and the fresh multi-seed values are
reported instead. Runs cost 2--6 minutes each on a free Colab GPU;
optimizer and solver settings are identical across pipelines
(AdamW, $3{\times}10^{-4}$, weight decay $10^{-4}$, global-norm clip
$1.0$).

\section{Conclusion}\label{sec:conclusion}
We set out to build a physics-structured implicit reasoner and instead
built a very careful copy machine. The equilibrium of our
port-Hamiltonian DEQ is its initialization --- under the anchored
design, without the anchor, and even when retrained from randomized
starts --- and the one training run that escaped this regime did so by
diverging into decoder-adapted noise. Nothing in ordinary training
telemetry reveals any of this; it took substitution tests,
gradient-path inspection, and controlled retraining arms to see it.

We draw three implications. First, for the growing literature that
credits looped, recurrent-depth, and implicit models with
\emph{reasoning}: the credit assumes the iterated module computes,
and that assumption is cheap to test and capable of failing silently
--- we suggest the substitution test (T1) as a standard reported
control. Second, for architecture design: when a task is solvable
through a direct pathway, adding an expensive iterative module does
not merely risk wasted parameters; the optimizer will actively route
around it, and stabilizing structure can make the do-nothing solution
easier to reach. Gradient incentives, not architectural affordances,
determine whether iteration is used. Third, for practice: negative
results of this shape are inexpensive to establish rigorously --- our
entire evidence base cost minutes per run on a free GPU --- and the
diagnostic protocol transfers directly to models far larger than
ours.

\paragraph{Reproducibility.} All experiments run on a single free
Colab GPU (per-run cost: $\sim$2--6 minutes training). Code, configs,
seeds, and raw logs: \url{https://github.com/joyjeet-singh/lazy-identity-deq}

\bibliographystyle{unsrtnat}
\bibliography{refs}

\appendix
\section{Why Physics Structure Seemed Worth Trying}\label{app:drift}
The port-Hamiltonian parameterization was not chosen arbitrarily. In a
standalone rollout experiment on learned two-body-style dynamics, we
compared long-horizon energy behavior of an unconstrained baseline
network against the structure--dissipation parameterization used in
this paper. Over a $5{,}000$-step rollout the baseline's normalized
Hamiltonian drift saturates at $6.2\times10^{25}$ ($3.5\times10^{17}$
already by step $501$), while the port-Hamiltonian network holds
$2.3\times10^{3}$ (and $2.3\times10^{-2}$ at step $501$) --- roughly
twenty orders of magnitude of improvement in energy control. This is
the standard, genuine benefit of the inductive bias, and it is
precisely why the main text's finding is worth reporting: stability of
the dynamics and usefulness of the equilibrium are different
properties, and the first does not purchase the second.

\section{Training Curves and Instability Window}\label{app:curves}
\begin{figure}[h]\centering
\includegraphics[width=.95\linewidth]{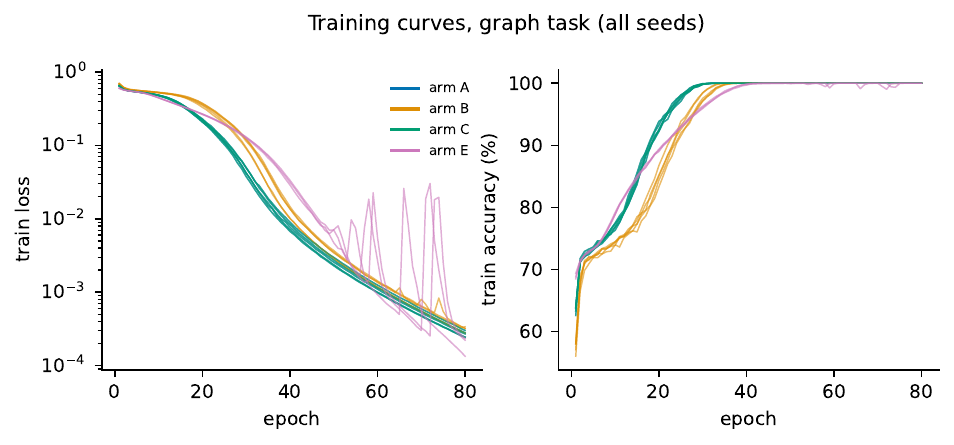}
\caption{Training curves, graph task, all arms and seeds.}
\label{fig:curves}
\end{figure}

Figure~\ref{fig:curves} shows training loss and accuracy for every
seed of every graph-task arm; all runs train to convergence
(100\% train accuracy), closing the under-training explanation. For
the ProofWriter gate-variant instability (Section~\ref{sec:gate}),
the instrumented onset window recorded per-epoch peak state magnitude
$\max|\hstar|$ rising $6.9\to21.7$ across epochs 74--90 with
initializer gradient spikes to $1.7$ and a $92\%\to62\%$ accuracy
excursion; solver mean residuals in this variant reached $29.1$
(cf.\ $0.09$ in the standard pipeline), i.e.\ returned states were
not equilibria.

\section{Full Experimental Details}\label{app:details}
\textbf{Optimization.} AdamW, learning rate $3\times10^{-4}$, weight
decay $10^{-4}$, global-norm clip $1.0$; batch size $64$;
$100$ epochs (ProofWriter) / $80$ epochs (graphs); ProofWriter uses an
auxiliary-weight ramp $\alpha = 0.1\min(1,\text{epoch}/20)$.
\textbf{Solver.} Anderson acceleration, memory $m{=}5$, iteration cap
$300$, Tikhonov-regularized mixing with coefficient clipping to
$[-3,3]$ and state clipping to $[-50,50]$; adjoint solve: cap $100$,
tolerance $10^{-4}$. \textbf{Model sizes.} Phase dimension $6$
(ProofWriter) / $8$ (graphs); context dimension $128$; kinetic and
potential MLPs with hidden width $32$ and softplus activations;
decoder hidden width $64$ (GELU). \textbf{Data.} ProofWriter examples
scanned up to a $60{,}000$-row cap and bucketed by proof depth to the
splits in Section~\ref{sec:setup}; graph datasets generated with a
fixed data seed ($42$), exactly class-balanced, with dataset SHA-256
recorded in every run manifest. \textbf{Hardware and cost.} Single
free-tier Colab GPU; $\sim$2 minutes per ProofWriter run,
$\sim$5--6 minutes per graph run ($\sim$26 at $30$k examples).
\textbf{Logging.} Every run writes a manifest (library versions,
device, config hash, dataset hash) and per-epoch JSONL records; all
tables and figures in this paper are generated from those files by the
released analysis scripts.
\end{document}